\documentclass{article}

\usepackage{arxiv}          
\usepackage[utf8]{inputenc} 
\usepackage[T1]{fontenc}    
\usepackage{hyperref}       
\usepackage{url}            
\usepackage{booktabs}       
\usepackage{amsfonts}       
\usepackage{nicefrac}       
\usepackage{microtype}      
\usepackage{graphicx}       
\usepackage{enumitem}       
\usepackage{tabularx}       
\usepackage{caption}        
\usepackage{multirow}       
\usepackage{soul}           
\usepackage{xcolor}         
\usepackage{array}          
\usepackage{amsmath}        
\usepackage{subcaption}     
\usepackage[style=numeric,sorting=none]{biblatex}

\date{} 
\renewcommand{\undertitle}{} 
\renewcommand{\headeright}{} 

\newenvironment{figure_in_table}{\captionsetup{type=figure}}{}
\newcolumntype{P}[1]{>{\centering\arraybackslash}p{#1}}

\hypersetup{
    colorlinks=true,
    urlcolor=blue,
    pdftitle={Homogeneous Vector Capsules Enable Adaptive Gradient Descent in Convolutional Neural Networks},
    pdfpagemode=FullScreen,
}

\title{Homogeneous Vector Capsules Enable Adaptive Gradient Descent in Convolutional Neural Networks}

\author{
  Adam~Byerly\\
  Department of Electronic and Electrical Engineering\\
  Brunel University London\\
  Uxbridge, UB8 3PH UK \\
  Department of Computer Science and Information Systems\\
  Bradley University\\
  Peoria, Il, 61615 USA\\
  \texttt{abyerly@fsmail.bradley.edu} \\
  \And{}
  Tatiana~Kalganova \\
  Department of Electronic and Electrical Engineering\\
  Brunel University London\\
  Uxbridge, UB8 3PH UK \\
  \texttt{tatiana.kalganova@brunel.ac.uk} \\
}

\begin{document}

\maketitle

\begin{abstract}
Capsules are the name Geoffrey Hinton has given to vector-valued neurons.  Neural networks traditionally produce a scalar value for an activated neuron.  Capsules, on the other hand, produce a vector of values, which Hinton argues correspond to a single, composite feature wherein the values of the components of the vectors indicate properties of the feature such as transformation or contrast.  We present a new way of parameterizing and training capsules that we refer to as homogeneous vector capsules (HVCs).  We demonstrate, experimentally, that altering a convolutional neural network (CNN) to use HVCs can achieve superior classification accuracy without increasing the number of parameters or operations in its architecture as compared to a CNN using a single final fully connected layer.  Additionally, the introduction of HVCs enables the use of adaptive gradient descent, reducing the dependence a model's achievable accuracy has on the finely tuned hyperparameters of a non-adaptive optimizer.  We demonstrate our method and results using two neural network architectures.  For the CNN architecture referred to as Inception v3, replacing the fully connected layers with HVCs increased the test accuracy by an average of 1.32\% across all experiments conducted.  For a simple monolithic CNN, we show HVCs improve test accuracy by an average of 19.16\%.
\end{abstract}

\keywords{Adaptive Gradient Descent, Capsule, Convolutional Neural Network (CNN), Homogeneous Vector Capsules (HVCs), Inception}

\section{Introduction}\label{sec:introduction}

In~\cite{Hinton2011}, the authors argued that standard convolutional neural networks are ``misguided'' in their usage of neurons that are composed of singular scalars to summarize their activation.  The authors proposed (a) the concept of a ``capsule'', which is comprised of multiple scalar values and (b) posited that these capsules would be capable of recognizing a ``visual entity over a limited domain of viewing conditions and deformations'' and that the capsule's members would include both the probability that the entity is present as well as a set of ``instantiation parameters'' that ``may include the precise pose, lighting and deformation relative to the canonical version of that entity''.  In their work, they (c) demonstrated that capsules could learn the \(x\) and \(y\) coordinates of a visual entity and (d) made a convincing case that capsules could learn to identify ``any property of an image that we can manipulate in a known way''.

Research into capsules did not progress much until a pair of papers were pre-published on arXiv in late 2017.  The first of these two papers (\cite{Sabour2017}) received an especially significant amount of attention, due to the fact that it published results on par with the state-of-the-art for both the standard MNIST~\cite{LeCun1998} and smallNORB~\cite{LeCun2004} datasets using a relatively shallow network in combination with capsules.  Additionally, the network described in the first paper was shown to be highly effective at segmenting highly overlapped digits from the MNIST data.  Both papers utilized an iterative routing mechanism between layers of capsules.  They referred to the method in the first paper as ``Dynamic Routing'' and used a different method in the second paper based on the Expectation-Maximization algorithm~\cite{Dempster1977}.  The architecture described in the second paper (\cite{Hinton2018}) improved upon the state-of-the-art classification accuracy for smallNORB by 45\%.

The architectures described in both papers used two layers of capsules in order to make the final classification and used matrix multiplication between them.  In both papers, in addition to learning the weights used in the matrix multiplications using backpropagation, a routing algorithm was employed to iteratively ``refine'' the weights of the matrices.  The authors interpret the first set of capsules as ``parts'' and the second set as ``wholes'' and the routing algorithm as a method for finding agreement about which whole is best described by the particular set of parts.

Both papers published results on relatively small data sets.  In both cases this was due to the high computational cost associated with using a routing algorithm.  Additionally, the architecture from the first paper requires a large number of parameters per output class (147,456) just for the weights between capsule layers, making datasets with a large number of output classes (like the 1,000 classes in ImageNet) intractable.

Another important thread of neural network research is choosing the best optimization algorithm and its hyperparameters.  Stochastic Gradient Descent (SGD) with momentum is simple and effective but requires careful tuning of both the learning rate \(\eta{}\) and the schedule for decaying that learning rate as training progresses.  Though guidance has emerged in the form of rules-of-thumb~\cite{Wilson2018}, it is none-the-less true that the choice of the learning rate and rate decay scheme remain a matter of trial-and-error and heavily dependent on the data being trained on.  As such, alleviating the need to carefully tune a single learning rate has emerged as an important research area.

The most successful strategy for alleviating the need to carefully tune the learning rate has been to maintain separate learning rates for every trainable parameter and to learn each of these learning rates based on the magnitude of previous gradient updates to those parameters.  This method in general is referred to as adaptive gradient descent.  Research into this began in earnest with AdaGrad~\cite{Duchi2011} and has continued to be an active area of research up to the present, with the most popular adaptive method currently being Adam~\cite{Kingma2014}.  Adaptive methods of gradient descent are popular for several reasons.  First, because they adapt a learning rate for every parameter, they are able to learn sparse, yet highly informative features differently than more dense information that may be less predictive.  Second, they reduce the need for careful tuning of the learning rate and learning rate decay by allowing the learning rate to be ``learned'' from the data.  And third, they tend to approach a convergence much earlier in the training scheme compared to non-adaptive methods for the same data and network.

Unfortunately, adaptive gradient descent methods have some weaknesses.  First, sparsely occurring features that are not highly informative have overweight influence relative to less sparsely occurring features.  And second, empirically, they are prone to overfitting and creating a generalization gap between the in-sample and out-of-sample predictions.  This has led some researchers to state that the generalization gap of adaptive gradient descent methods is an open problem~\cite{Chen2018a} and has led other researchers to recommend not using adaptive methods at all~\cite{Wilson2018}.  Indeed, the best performing convolutional neural networks (CNNs) of the past few years have all used non-adaptive gradient descent methods and hand-tuned learning rate decay schemes~\cite{Simonyan2015}\cite{Szegedy2015a}\cite{Szegedy2015b}\cite{He2015}\cite{Szegedy2016}\cite{Chollet2017}.

Our contribution is as follows:
\begin{enumerate}
  \item We present a new way of parameterizing and training a pair of capsule layers which we call homogeneous vector capsules (HVCs).  This method, as compared to other prevailing capsule methods (see~\cite{Sabour2017},~\cite{Hinton2018},~\cite{Venkataraman2020}, and~\cite{Amer2020}), uses drastically fewer parameters and avoids expensive iterative routing procedures, instead relying solely on weights learned during backpropagation.
  \item We demonstrate experimentally that classifying with HVCs, rather than classifying with fully connected layers, achieves massively superior results in a simple monolithic CNN and quantifiably superior results in a more advanced CNN architecture (Inception v3).
  \item We show that the practice of using large values of \(\epsilon{}\) in popular adaptive gradient descent methods has the effect of muting their adaptability.
  \item We demonstrate experimentally that when using HVCs, training with adaptive gradient descent methods using the intended small value of \(\epsilon{}\) restores the adaptability of the methods and achieves superior classification accuracy relative to finely-tuned learning rates and decay schedules.  In so doing, we solve an open problem in convolutional neural network research.
\end{enumerate}

\section{Related Work}\label{sec:related_work}

Morzhakov et al.~\cite{Morzhakov2017}, inspired by the work of Hubel \& Wiesel~\cite{Hubel1968}, put forth a neural network architecture similar to that used by~\cite{Sabour2017} in that it utilized vector neurons, rather than scalar neurons, which shared common inputs and outputs.  As their work was inspired by the physiology of primate brains, they characterized the structure as minicolumns, the term used for the analogous structure in primate brains.  It is noteworthy that their architecture did not use any analog to the routing mechanism employed by~\cite{Sabour2017} and~\cite{Hinton2018}.  While performing comparably with traditional CNNs on the MNIST dataset, it performed worse than the architecture employed by~\cite{Sabour2017}.

Roy et al.~\cite{Roy2018}, compared the effects of various forms of image degradation (additive white gaussian noise, salt and pepper noise, etc.) on MobileNet~\cite{Howard2017}, VGG16 \& VGG19~\cite{Simonyan2015}, Inception v3~\cite{Szegedy2015b}, and CapsNet~\cite{Sabour2017} and found that CapsNet was far more robust against the degradation methods they tested than any of the others.  They hypothesize that this is not only due to the presence of the capsule neurons and/or dynamic routing, but also due to the shallower nature of CapsNet, having gone through fewer layers of convolutions.

Nair et al.~\cite{Nair2018}, ventured to apply the CapsNet architecture proposed by~\cite{Sabour2017} to more complex datasets than MNIST---Fashion MNIST~\cite{Xiao2017}, SVHN~\cite{Netzer2011}, and CIFAR-10~\cite{Krizhevsky2009}.  Additionally, they experimented with a greater range of affine deformations than the small amount of translation used in the original experiments.  Their conclusion was that the CapsNet architecture is ``unlikely to work on other classification tasks, let alone machine learning tasks in general''.  They also concluded that the design was ``not making full use of routing to encode'' the spatial relationships between the components of the objects the network was classifying.  They hypothesized that a neural network, as opposed to a routing algorithm, would better accomplish the goal of reweighting the coefficients used to determine the agreement between capsule layers.  This method was experimented with by~\cite{Chen2018b}, though they were unable to produce any significant results.  Additionally, they hypothesized that for data more complex than MNIST, deeper networks may be required.  We agree with these last two hypotheses and for our experiments, (1) we use a neural network approach, rather than a routing approach, when transforming between capsule layers, and (2) we use deeper networks for classifying image data that is much more complex than MNIST.\@

Fang et al.~\cite{Fang2018} applied a capsule network to the task of protein gamma-turn prediction, rather than to image classification---the first such application of capsule networks in the bioinformatics domain.  Novel to their experiments is that they prepended the capsules portion of the network with an inception block ala Szegedy et al.~\cite{Szegedy2015b} rather than a simple convolution.  They achieved a new state-of-the-art performance on the GT320 benchmark~\cite{Guruprasad2000} for gamma-turn prediction with an MCC (Matthew correlation coefficients---the metric used for this task) of 0.45, beating the previous state-of-the-art of 0.38.

One of the best performing CNNs to be published in the past several years is Inception v3~\cite{Szegedy2015b}.  To train their architecture, they used the RMSProp\footnote{RMSProp is an unpublished, adaptive learning rate method introduced by Geoffrey Hinton in Lecture 6e of a now no longer available Coursera course.  See: \href{http://www.cs.toronto.edu/~tijmen/csc321/slides/lecture_slides_lec6.pdf}{http://www.cs.toronto.edu/\textasciitilde{}tijmen/csc321/slides/lecture\textunderscore{}slides\textunderscore{}lec6pdf}} optimizer, which is indeed designed to be an adaptive gradient descent method.  RMSProp adapts each parameter in the model using \autoref{equation:equation1}:

\begin{equation}
  \frac{1}{\sqrt{E[g^{2}]+ \epsilon}}
  \label{equation:equation1}
\end{equation}

\(E[g^2]\) is the exponential moving average of the past squared gradients for the parameter and the intended purpose of the \(\epsilon{}\) parameter is to provide numeric stability by mitigating the danger of division by zero, and thus implementations default this value to \(1\times{}10^{-10}\), which would create a range of possible values for the per-parameter adaptive term of \(0\) to \(1\times{}10^{6}\).  By using a value of 1.0 when training Inception v3, they limit this range to 0 to 1, thus setting an upper bound five orders of magnitude less than intended for this term.  While still technically adapting each parameter, the range of adaptation is so dampened that we would characterize RMSProp with a 1.0 \(\epsilon{}\) as quasi-adaptive at best.  As such, we agree with Chen and Gu~\cite{Chen2018a} that effectively utilizing (truly) adaptive gradient descent methods with convolutional neural networks remains an open problem relative to Inception v3.

The Adam optimizer~\cite{Kingma2014} has an analogous per-parameter adaptive term for each of the past squared gradients shown in \autoref{equation:equation2} (in addition to another term not relevant to this discussion for past gradients that gives Adam a momentum-like behavior):

\begin{equation}
  \frac{1}{\sqrt{\hat{v}_t} + \epsilon}
  \label{equation:equation2}
\end{equation}

\(\hat{v}_t\) is the bias corrected exponential moving average of the past squared gradients for the parameter.  Here again, the Adam optimizer employs the use of an \(\epsilon{}\) that implementations default to \(1\times{}10^{-10}\).  Since in Adam, the \(\epsilon{}\) is moved out from underneath the radical, Adam is able to adapt each parameter by five orders of magnitude more than RMSProp (with a range of \(0\) to \(1\times{}10^{10}\)).

\section{Capsule Layers Configuration}\label{sec:capsule_layers_configuration}

Sabour et al.~\cite{Sabour2017} proposed two final layers of capsules.  The first of which has 8 dimensions shaped as a vector and the second of which has 16 dimensions, also shaped as a vector.  The transformation between the two layers of capsules is a typical matrix multiplication, wherein every pair of capsules has an associated 16\(\times{}\)8 matrix of trainable parameters and is multiplied by each of the 8-dimensional vector capsules and summed to form the input into the 16-dimensional capsule.  In \autoref{equation:equation3}, an equivalent transformation simplified to two and four dimensions for clarity is presented.

\begin{equation}
  \begin{bmatrix}
    a & b \\
    c & d \\
    e & f \\
    g & h
  \end{bmatrix}
  \cdot
  \begin{bmatrix}
    x_1 \\
    x_2
  \end{bmatrix}
    =
  \begin{bmatrix}
    ax_1 + bx_2 \\
    cx_1 + dx_2 \\
    ex_1 + fx_2 \\
    gx_1 + hx_2
  \end{bmatrix}
  \label{equation:equation3}
\end{equation}

In this equation, as well as in \autoref{equation:equation4} and \autoref{equation:equation5}, the variables \(a\) through \(h\) represent learned weights that are being applied in the transformation, and the variables \(x_1\) through \(x_2\) (in \autoref{equation:equation3}) and \(x_4\) (in \autoref{equation:equation4} and \autoref{equation:equation5}) represent the values computed from the previous operation in the network.  In each case, the second matrix on the left hand side of the equation is the first capsule and the matrix on the right hand side is the second capsule.

A problem with this transformation in \autoref{equation:equation3} becomes apparent when viewing it as an overdetermined system of linear equations in matrix form: every dimension in the second layer of capsules, beyond the dimensions in the first layer, are at best redundant and more probably, due to the random initialization of the weights, a challenge to the optimization algorithm used during backpropagation to reconcile multiple differing losses derived from each activation in the previous layer.

Also, it should be noted that each dimension of the second layer of capsules is a linear combination of all dimensions of the first layer of capsules.  This is a desirable property in a fully connected layer in a neural network.  However, with the interpretation and empirical verification in the work of Sabour et al.~\cite{Sabour2017} of the dimensions of a capsule as being distinct features of a given sample, it is our hypothesis that this entangling of distinct features from one layer into all features in the next layer is an undesirable property.

In their follow-up work, Hinton et al.~\cite{Hinton2018} switched to using an equivalent number of dimensions in neighboring capsule layers, though they did not cite their motivation for doing so as to alleviate the problem of an overdetermined system.  Additionally, they shaped their capsules as matrices rather than vectors.  The authors noted that this reshaping had the effect of reducing the number of trainable parameters (for every pair of capsules) from being the product of the dimensions of the two layers of capsules to being only the number of dimensions of a single layer of capsules.  This method of matrix capsules requires that the number of dimensions in neighboring layers be both equivalent and a perfect square.  In \autoref{equation:equation4}, an equivalent transformation simplified to four dimensions is presented:

\begin{equation}
  \begin{bmatrix}
    a & b \\
    c & d
  \end{bmatrix}
  \cdot
  \begin{bmatrix}
    x_1 & x_2 \\
    x_3 & x_4
  \end{bmatrix}
    =
  \begin{bmatrix}
    ax_1 + bx_3 & ax_2 + bx_4 \\
    cx_1 + dx_3 & cx_2 + dx_4
  \end{bmatrix}
  \label{equation:equation4}
\end{equation}

In addition to alleviating the problem of an overdetermined system and significantly reducing the number of trainable parameters, this formulation results in only the square root of the total number of features in the first layer being entangled with each feature in the second layer.

We propose a new method for the transformation from one layer of capsules to the next.  Rather than using the typical transformation matrix, the proposed method involves using a transformation vector and rather than using the typical matrix multiplication, the proposed method involves using the Hadamard product (element-wise multiplication).  This method is shown in \autoref{equation:equation5}, simplified to four dimensions for clarity:

\begin{equation}
  \begin{bmatrix}
    a \\
    b \\
    c \\
    d
  \end{bmatrix}
  \odot
  \begin{bmatrix}
    x_1 \\
    x_2 \\
    x_3 \\
    x_4
  \end{bmatrix}
    =
  \begin{bmatrix}
    ax_1 \\
    bx_2 \\
    cx_3 \\
    dx_4
  \end{bmatrix}
  \label{equation:equation5}
\end{equation}

This method goes back to using vectors for the shape of the capsules and requires that the neighboring layers of capsules be of equivalent dimension, thus we call these homogeneous vector capsules.  With the constraint of requiring equivalent dimensions in the capsule layers, this method comes with the following benefits:
\begin{enumerate}
\item Because this method uses the Hadamard product rather than typical matrix multiplication, the drawback of using the more intuitive vector shape for a capsule is removed, as the number of trainable parameters per pair of capsules stays equal to the number of dimensions in the capsules (as in Hinton et al.~\cite{Hinton2018}), rather than being that number of dimensions squared (as in Sabour et al.~\cite{Sabour2017}).
\item By the nature of the Hadamard product, this method cannot suffer from the problem of an overdetermined system.
\item This fully disentangles features from the dimensions in the first layer of capsules from differing dimensions in the subsequent layer of capsules.\@i.e., each dimension in the first layer maps to one and only one dimension in the second layer.
\item This eliminates all of the addition operations used in matrix multiplication for a modest reduction in computational cost.
\item Whereas the number of dimensions in~\cite{Hinton2018} must be a perfect square, HVCs can be composed of any number of dimensions that evenly divides the number of neurons being input into them.
\end{enumerate}

\section{Experimental Setup and Results}\label{sec:experimental_setup_and_results}

We designed our experiments to compare (a) baseline neural network architectures that use the standard approach of transforming the final convolutional layer in the network as in \autoref{fig:fully_connected_before_output} with (b) reshaping the final set of feature maps into \(j\) \(n\)-dimensional vector capsules, where \(j\cdot n\) is the total number of weights coming out of the final set of feature maps.  When doing this, the final classification is done, rather than with scalar output neurons, with \(y\) \(n\)-dimensional vector capsules as in \autoref{fig:capsules_before_output}, that are reduced to predictions by computing the Euclidian norm of the vectors.

\setlength\tabcolsep{0.02in}
\begin{tabular}{@{}lp{3.11in}llp{3.11in}l@{}}
  \multicolumn{3}{l}{\includegraphics[width=3.21in]{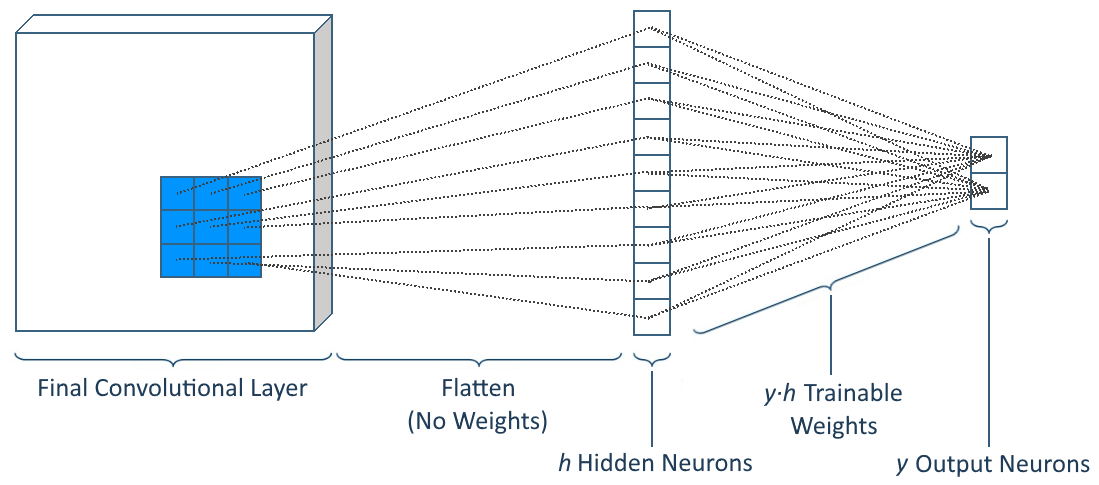}} &
  \multicolumn{3}{l}{\includegraphics[width=3.21in]{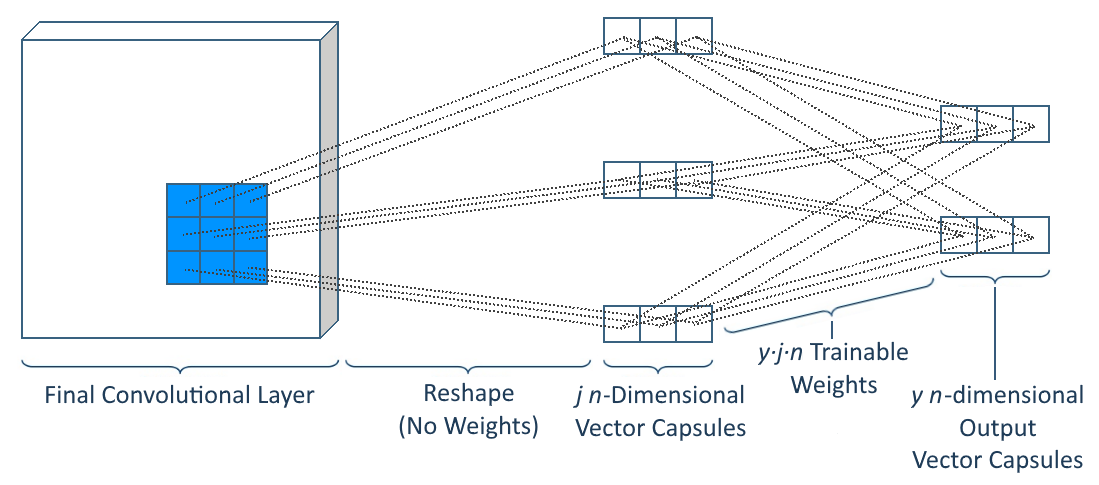}} \\[-0.05in]
  &
  \begin{figure_in_table}
    \caption{The standard approach to transforming the final convolutional layer into class predictions.}\label{fig:fully_connected_before_output}
  \end{figure_in_table} &
  &
  &
  \begin{figure_in_table}
    \caption{Using homogeneous vector capsules to transform the final convolutional layer into class predictions.}\label{fig:capsules_before_output}
  \end{figure_in_table} &
\end{tabular}

We conducted our experiments using two convolutional neural network architectures.  The first network is a typical simple monolithic CNN featuring a series of 3\(\times{}\)3 convolutions interspersed with max pooling operations (see \autoref{tab:simple_monolithic_cnn_ops}).  The motivation behind this design was to examine the effect of capsules on a simple, widely understood and easily implemented architecture with a low number of parameters (in this case \textasciitilde{} 1.6M to \textasciitilde{} 22.1M, depending on the number of output classes and the capsule configuration).  We used no drop-out or L2 (or any other form) of regularization with this architecture.  The second network is the popular Inception v3 architecture~\cite{Szegedy2015b}.  This network was chosen due to its good performance given the relatively low number of parameters it uses (\textasciitilde{} 23.2M to \textasciitilde{} 156.1M, depending on the number of output classes and the capsule configuration).  

\setlength\tabcolsep{6pt}
\begin{table}[!htbp]
  \caption{The stem of the simple monolithic CNN.\@  The baseline experiments were classified through a fully connected layer after flattening the final set of feature maps as in \autoref{fig:fully_connected_before_output}.  All other experiments used HVCs as in \autoref{fig:capsules_before_output}.}
  \begin{tabularx}{\textwidth}{@{}Xrr@{}}
    \toprule
      Operation & Feature Maps & Output Shape\\
    \midrule
      3\(\times{}\)3 convolution w/stride 2 & 32   & 149\(\times{}\)149\(\times{}\)32\\
      3\(\times{}\)3 convolution w/stride 1 & 32   & 147\(\times{}\)147\(\times{}\)32\\
      3\(\times{}\)3 convolution w/stride 1 & 32   & 145\(\times{}\)145\(\times{}\)32\\
      2\(\times{}\)2 max pool w/stride    1 & N/A* & 72\(\times{}\)72\(\times{}\)32\\
      3\(\times{}\)3 convolution w/stride 1 & 64   & 70\(\times{}\)70\(\times{}\)64\\
      3\(\times{}\)3 convolution w/stride 1 & 64   & 68\(\times{}\)68\(\times{}\)64\\
      3\(\times{}\)3 convolution w/stride 1 & 64   & 66\(\times{}\)66\(\times{}\)64\\
      2\(\times{}\)2 max pool w/stride    1 & N/A* & 33\(\times{}\)33\(\times{}\)64\\
      3\(\times{}\)3 convolution w/stride   & 128  & 31\(\times{}\)31\(\times{}\)128\\
      3\(\times{}\)3 convolution w/stride 1 & 128  & 29\(\times{}\)29\(\times{}\)128\\
      3\(\times{}\)3 convolution w/stride 1 & 128  & 27\(\times{}\)27\(\times{}\)128\\
      2\(\times{}\)2 max pool w/stride    1 & N/A* & 13\(\times{}\)13\(\times{}\)128\\
      3\(\times{}\)3 convolution w/stride 1 & 256  & 11\(\times{}\)11\(\times{}\)256\\
      3\(\times{}\)3 convolution w/stride 1 & 256  & 9\(\times{}\)9\(\times{}\)256\\
    \bottomrule
  \end{tabularx}\\
  \captionsetup{width=\textwidth}
  \caption*{* Max-pooling is a sub-sampling operation that involves no trainable parameters.}\label{tab:simple_monolithic_cnn_ops}
\end{table}

We conducted our experiments using three datasets of increasing difficulty:

\begin{itemize}
\item The full-sized Imagenette~\cite{imagenette}, a subset of ImageNet consisting of 10 easily classified classes: tench, English springer, cassette player, chain saw, church, French horn, garbage truck, gas pump, golf ball, and parachute.
\item The full-sized Imagewoof~\cite{imagenette}, a subset of ImageNet consisting of 10 more closely related classes, all of which are dog breeds: Australian terrier, Border terrier, Samoyed, Beagle, Shih-Tzu, English foxhound, Rhodesian ridgeback, Dingo, Golden retriever, and Old English sheepdog.
\item Food-101~\cite{bossard14}, a challenging and noisy dataset consisting of 101 classes of images retrieved from the now defunct foodspotting.com.
\end{itemize}

We conducted our experiments using four different optimization strategies.  RMSProp has been a popular choice for optimizing convolutional neural networks since~\cite{Szegedy2015b}.  This strategy which we have denoted O1 (see \autoref{tab:optimizers}) is the strategy used in~\cite{Szegedy2015b} whereas the strategy we have denoted O2 is the strategy employed by the official TensorFlow implementation of Inception v3 published on github.com\footnote{\href{https://github.com/tensorflow/models/blob/master/research/slim/train_image_classifier.py}{https://github.com/tensorflow/models/blob/master/research/slim/train\textunderscore{}image\textunderscore{}classifier.py}} which results in slightly higher accuracy.  In addition, we experimented with two other optimization strategies.  O3 is the Adam optimizer with the defaults suggested in~\cite{Kingma2014}, and O4 is the Adam optimizer with a slowly decaying base learning rate.

\setlength\tabcolsep{6pt}
\begin{table}[!htbp]
  \captionsetup{width=\textwidth}
  \caption{Optimizers used for all experiments.}
  \begin{tabularx}{\textwidth}{@{}P{0.75in}X@{}}
    \toprule
      Optimizer \# & Description \\
    \midrule
      O1 & RMSProp w/epsilon 1 and 0.045 learning rate exponentially decaying every 2 epochs by 0.94 \\
      O2 & RMSProp w/epsilon 1 and 0.1 learning rate exponentially decaying every 30 epochs by 0.16\\
      O3 & Adam w/0.001 learning rate \\
      O4 & Adam w/0.001 learning rate exponentially decaying every epoch by 0.96 \\
    \bottomrule
  \end{tabularx}\label{tab:optimizers}
\end{table}

Additional experimental parameters are as follows:

\begin{itemize}
\item All activations were ReLU preceded by batch normalization~\cite{Ioffe2015}.
\item Loss for the Inception v3 experiments was computed using the label-smoothing regularization method as in~\cite{Szegedy2015b}, whereas categorical cross-entropy was used for the simple monolithic CNN experiments.
\item All experiments ran for 100 epochs.
\item Evaluations were performed using the exponential moving average of past weights as in~\cite{Izmailov2018}, with a decay factor of 0.999.
\item A batch size of 32 was used for Imagenette and Imagewoof. A batch size of 96 was used for Food-101 for models S1-S8 and I1-I6 and a batch size of 68 for models I7 and I8 (see \autoref{tab:simple_models} and \autoref{tab:inception_v3_models}).  These batch sizes were dictated by the constraints of the available hardware.
\item We used an image size of 299\(\times{}\)299 for all images in all datasets, in all cases augmented using the strategy employed by the official TensorFlow implementation of Inception v3 published on github.com.\footnote{\href{https://github.com/tensorflow/models/blob/master/research/slim/preprocessing/inception_preprocessing.py}{https://github.com/tensorflow/models/blob/master/research/slim/preprocessing/inception\textunderscore{}preprocessing.py}}
\end{itemize}

\begin{figure}[!htbp]
  \subfloat[In this example, the filter maps have been converted into two 2-dimensional capsules for each distinct \(x\) and \(y\) coordinate of the feature maps.  The first 2 of 18 such capsules are highlighted in red and blue respectively.  Models S2, I2, I3, I4, I5, and the main output of I6 use this method.]{
  \begin{minipage}[c][1\width]{0.3\textwidth}
     \centering
     \includegraphics[width=0.6\textwidth]{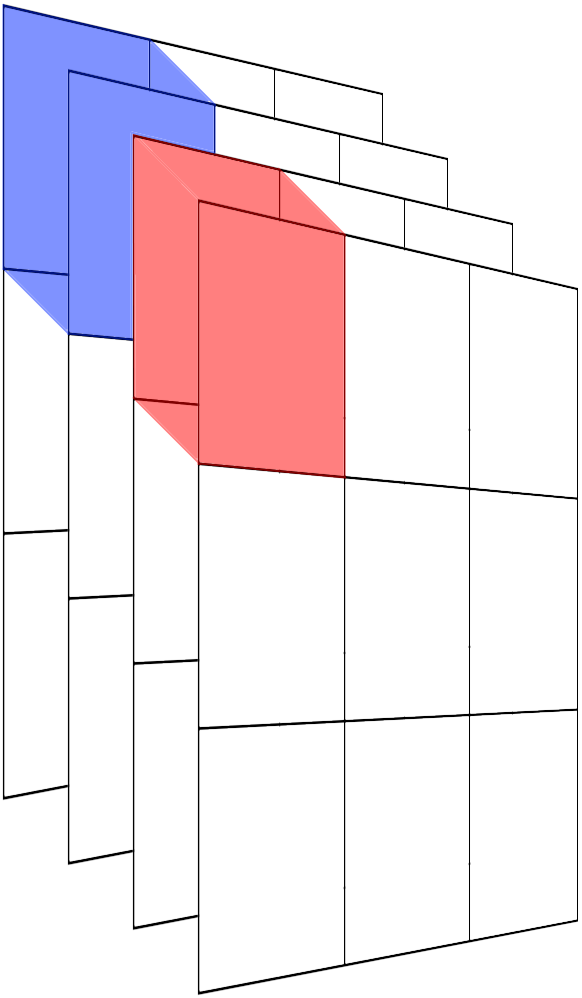}\label{fig:capsule_method_a}
  \end{minipage}}
  \hfill
  \subfloat[In this example, the filter maps have been converted into a single 4-dimensional capsule for each distinct \(x\) and \(y\) coordinate of the feature maps.  The first 2 of 9 such capsules are highlighted in red and blue respectively.  Models S4, I8, and the auxiliary output of I6 use this method.]{
  \begin{minipage}[c][1\width]{0.3\textwidth}
    \centering
    \includegraphics[width=0.6\textwidth]{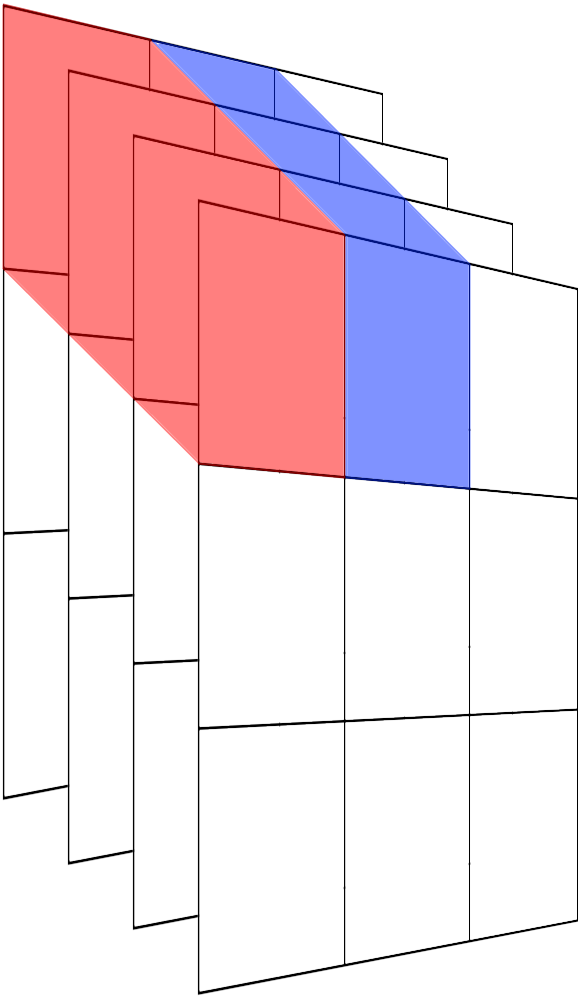}\label{fig:capsule_method_b}
  \end{minipage}}
  \hfill  
  \subfloat[In this example, the filter maps have been converted into four 9-dimensional capsules, each made from an entire feature map.  The first 2 of 4 such capsules are highlighted in red and blue respectively.  Models S3 and I7 use this method.]{
  \begin{minipage}[c][1\width]{0.3\textwidth}
    \centering
    \includegraphics[width=0.6\textwidth]{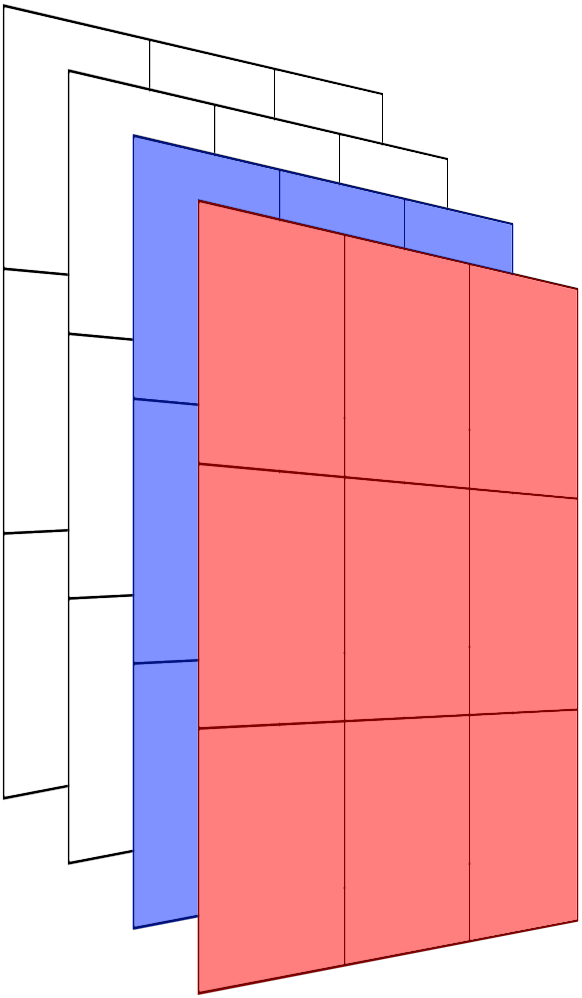}\label{fig:capsule_method_c}
  \end{minipage}}
  \caption{Methods of converting a set of four 3\(\times{}\)3 filter maps into a set of capsules.}\label{fig:capsule_methods}
\end{figure}

In our experiments, we explored 3 different methods of transforming the final set of feature maps into capsules.  The first method creates multiple capsules for each distinct \(x\) and \(y\) coordinate of the feature maps (see \autoref{fig:capsule_method_a}).  The second method creates a single capsule for each distinct \(x\) and \(y\) coordinate of the feature maps (see \autoref{fig:capsule_method_b}).  The intuition behind these two methods is that each position in the feature map represents a meaningful feature and that using capsules to ``group'' these together from multiple filter maps encourages the feature maps to cooperate.  The difference being that in the multiple capsule case, multiple disparate groups are allowed, wherein the single capsule case, only one such group is allowed.  The third method creates a single capsule for each distinct feature map (see \autoref{fig:capsule_method_c}).  The intuition behind this method is the standard interpretation of a feature map (i.e.\ it represents a single feature per map).  However, rather than allowing each dimension of the feature map to learn independently through a fully connected layer, we use capsules to maintain the cohesion among the dimensions.

We conducted experiments on eight variations of the simple monolithic CNN architecture (see \autoref{tab:simple_models}).  The first such variation, denoted S1, is the baseline model that flattens the final set of feature maps and then classifies through a layer of fully connected neurons.  Variations S2 through S6 reshape the final set of feature maps as in \autoref{fig:capsule_method_a}.  Variation S7 reshapes the final set of feature maps as in \autoref{fig:capsule_method_b}.  Variation S8 reshapes the final set of feature maps as in \autoref{fig:capsule_method_c}.  In variations S2 through S8, after the first layer of capsules are shaped, they are then classified through the second set of capsules that form the HVC pairs.

\setlength\tabcolsep{6pt}
\begin{table}[!htbp]
  \captionsetup{width=\textwidth}
  \caption{Models used for the experiments conducted using a simple monolithic CNN.}
  \begin{tabularx}{\textwidth}{@{}P{0.5in}Xrp{0.1in}r@{}}
    \toprule
      Model & Capsule Configuration & HVC Dimensions & & \# of HVCs \\
    \midrule
      S1 & \multicolumn{3}{l}{No capsules --- This is the baseline model} \\
      S2 & See \autoref{fig:capsule_method_a} & 8 & & 2,592 \\
      S3 & See \autoref{fig:capsule_method_a} & 16 & & 1,296 \\
      S4 & See \autoref{fig:capsule_method_a} & 32 & & 648 \\
      S5 & See \autoref{fig:capsule_method_a} & 64 & & 324 \\
      S6 & See \autoref{fig:capsule_method_a} & 128 & & 162 \\
      S7 & See \autoref{fig:capsule_method_b} & 256 & & 81 \\
      S8 & See \autoref{fig:capsule_method_c} & 81 & & 256 \\
    \bottomrule
  \end{tabularx}\\[-0.05in]\label{tab:simple_models} 
\end{table}

We conducted experiments on eight variations of the Inception v3 architecture (see \autoref{tab:inception_v3_models}).  The first such variation, denoted I1, is the baseline model as described in~\cite{Szegedy2015b}.  Variations I2 through I6 reshape the final set of feature maps in both the main and auxiliary branches as in \autoref{fig:capsule_method_a}.  Variation I7 reshapes the final set of feature maps in both branches as in \autoref{fig:capsule_method_b}, and Variation I8 reshapes the final set of feature maps in both branches as in \autoref{fig:capsule_method_c}.  In variations I2 through I8, after the first layer of capsules are shaped each branch is then classified through the second set of capsules that form the HVC pairs.

\setlength\tabcolsep{6pt}
\begin{table}[!htbp]
  \captionsetup{width=\textwidth}
  \caption{Models used for the experiments conducted using the Inception v3 architecture.}
  \begin{tabularx}{\textwidth}{@{}P{0.5in}Xrrp{0.1in}rr@{}}
    \toprule
      & &
      \multicolumn{2}{c}{{\small\underline{Main Out HVCs}}} & &
      \multicolumn{2}{c}{{\small\underline{Aux Out HVCs}}} \\
      Model & Capsule Configuration &
      Dimensions & \# & &
      Dimensions & \# \\
    \midrule
      I1 & \multicolumn{5}{l}{No capsules --- This is the baseline model} \\
      I2 & See \autoref{fig:capsule_method_a} & 8 & 256 & & 8 & 16 \\
      I3 & See \autoref{fig:capsule_method_a} & 16 & 128 & & 16 & 8 \\
      I4 & See \autoref{fig:capsule_method_a} & 32 & 64 & & 32 & 4 \\
      I5 & See \autoref{fig:capsule_method_a} & 64 & 32 & & 64 & 2 \\
      I6 & See \autoref{fig:capsule_method_a} & 128 & 16 & & 128 & 1 \\
      I7 & See \autoref{fig:capsule_method_b} & 2,048 & 64 & & 128 & 25 \\
      I8 & See \autoref{fig:capsule_method_c} & 64 & 2,048 & & 25 & 128 \\
    \bottomrule
  \end{tabularx}\\[-0.05in]\label{tab:inception_v3_models}
\end{table}

Unique to the Inception v3 architecture relative to the simple monolithic CNN is that, in the baseline model I1 and models I2 through I6, the final operation before the flattening operations in both the main and auxiliary outputs reduce the feature maps to 1\(\times{}\)1.  In the main branch, this is accomplished via global average pooling~\cite{Lin2014} and in the auxiliary branch, this is accomplished by performing a 5\(\times{}\)5 convolution on a set of 5\(\times{}\)5 feature maps.  Both of these methods effectively collapse the spatial information present in the preceding operations into a single scalar value per feature map.  Despite this, these global operations have been empirically shown to be effective in maintaining models' ability to achieve good generalization and accuracy, all while significantly reducing the number of trainable parameters.  Generally, these global operations precede a final fully connected layer from which classification is performed.  The larger the number of classes being classified, the more pronounced the reduction in trainable parameters is.  For two of our Inception v3 experiments, I7 and I8, we removed these global operations, which results in the final set of feature maps in the main branch being 8\(\times{}\)8 and the final set in the auxiliary branch being 5\(\times{}\)5.  This in turn results in an increasing number of parameters in the model as the number of output classes increases (see \autoref{tab:model_parameter_counts}).

\setlength\tabcolsep{6pt}
\begin{table}[!htbp]
  \captionsetup{width=\textwidth}
  \caption{Trainable parameters used by the models in our experiments by number of classes.  The difference between the number of trainable parameters for otherwise equivalent models using fully connected layers vs.\ HVCs is negligible.  For S1 vs.\ any of S2 through S8, the former has only 0.16\% fewer parameters.  As models get larger the difference lessens and eventually inverts.  For example, the difference between I1 and any of I2 through I6 when classifying 1000 classes is 0.007\% fewer parameters for the HVC models.}
  \begin{tabularx}{\textwidth}{ccXrr}
    \toprule
      Models & Final Feature Map Dimensions & & Classes & \# of Parameters  \\
    \midrule
      \multirow{3}{*}{\shortstack{S1-S8\\(see \autoref{tab:simple_models} and \autoref{tab:simple_dataset_optimizer_table})}} &
      \multirow{3}{*}{9\(\times{}\)9} &
      & 10 & 1.6M \\
      & & & 101 & 3.5M \\
      & & & 1000 & 22.1M \\
    \midrule
      \multirow{3}{*}{\shortstack{I1-I6\\(see \autoref{tab:inception_v3_models} and \autoref{tab:inception_v3_dataset_optimizer_table})}} &
      \multirow{3}{*}{1\(\times{}\)1} &
      & 10 & 22.3M \\
      & & & 101 & 22.5M \\
      & & & 1000 & 24.5M \\
    \midrule
      \multirow{3}{*}{\shortstack{I7-I8\\(see \autoref{tab:inception_v3_models} and \autoref{tab:inception_v3_dataset_optimizer_table})}} &
      \multirow{3}{*}{\shortstack{8\(\times{}\)8\\5\(\times{}\)5}} &
      & 10 & 23.2M \\
      & & & 101 & 35.4M \\
      & & & 1000 & 156.1M \\
    \bottomrule
  \end{tabularx}\label{tab:model_parameter_counts}
\end{table}

\clearpage

{\renewcommand{\arraystretch}{1.4}%
  \setlength\tabcolsep{3pt}
  \begin{table}[!htbp]
    \caption{Classification accuracy on 8 simple monolithic CNN models, 3 different datasets, and 4 different optimization strategies.  The first column in each row is a cross-referenc to the charts in \autoref{fig:simple_dataset_optimizer_charts}.  The model parentheticals refer to the number of capsule dimensions.  For example S2 (8d) refers to model S2 which uses 8-dimensional capsules.}
    \begin{tabularx}{\textwidth}{|X|c|c|P{0.47in}|P{0.47in}|P{0.47in}|P{0.47in}|P{0.47in}|P{0.47in}|P{0.55in}|P{0.55in}@{}|}
      \hline
        \multirow{2}{*}{Dataset} &
        \multirow{1}{*}{{\small \autoref{fig:simple_dataset_optimizer_charts}}} &
        \multirow{2}{*}{Optimizer} &
        \multicolumn{8}{c|}{Models (see \autoref{tab:simple_models})} \\
        \cline{4-11} & Chart & &
        \multicolumn{1}{c|}{S1} &
        \multicolumn{1}{c|}{S2 (8d)} &
        \multicolumn{1}{c|}{S3 (16d)} &
        \multicolumn{1}{c|}{S4 (32d)} &
        \multicolumn{1}{c|}{S5 (64d)} &
        \multicolumn{1}{c|}{S6 (128d)} &
        \multicolumn{1}{c|}{S7 (256d)} &
        \multicolumn{1}{c|}{S8 (81d)} \\
      \hline
        \multirow{4}{*}{Imagenette} & 1a & O1 &
        \multicolumn{1}{r|}{82.99\%} &
        \multicolumn{1}{r|}{85.02\%} &
        \multicolumn{1}{r|}{84.22\%} &
        \multicolumn{1}{r|}{85.43\%} &
        \multicolumn{1}{r|}{\textbf{87.58\%}} &
        \multicolumn{1}{r|}{86.91\%} &
        \multicolumn{1}{r|}{85.09\%} &
        \multicolumn{1}{r|}{86.24\%} \\
        & 1b & O2 &
        \multicolumn{1}{r|}{9.68\%} &
        \multicolumn{1}{r|}{85.66\%} &
        \multicolumn{1}{r|}{84.38\%} &
        \multicolumn{1}{r|}{86.68\%} &
        \multicolumn{1}{r|}{85.86\%} &
        \multicolumn{1}{r|}{85.71\%} &
        \multicolumn{1}{r|}{\textbf{88.65\%}} &
        \multicolumn{1}{r|}{87.12\%} \\
        & 1c & O3 &
        \multicolumn{1}{r|}{81.15\%} &
        \multicolumn{1}{r|}{85.78\%} &
        \multicolumn{1}{r|}{85.81\%} &
        \multicolumn{1}{r|}{85.76\%} &
        \multicolumn{1}{r|}{87.12\%} &
        \multicolumn{1}{r|}{\textbf{87.65\%}} &
        \multicolumn{1}{r|}{86.14\%} &
        \multicolumn{1}{r|}{82.30\%} \\
        & 1d & O4 &
        \multicolumn{1}{r|}{86.96\%} &
        \multicolumn{1}{r|}{88.11\%} &
        \multicolumn{1}{r|}{88.50\%} &
        \multicolumn{1}{r|}{88.91\%} &
        \multicolumn{1}{r|}{89.14\%} &
        \multicolumn{1}{r|}{89.32\%} &
        \multicolumn{1}{r|}{\textbf{89.63\%}} &
        \multicolumn{1}{r|}{88.42\%} \\
      \hline
        \multirow{4}{*}{Imagewoof} & 1e & O1 &
        \multicolumn{1}{r|}{21.49\%} &
        \multicolumn{1}{r|}{73.39\%} &
        \multicolumn{1}{r|}{76.00\%} &
        \multicolumn{1}{r|}{75.18\%} &
        \multicolumn{1}{r|}{78.43\%} &
        \multicolumn{1}{r|}{76.31\%} &
        \multicolumn{1}{r|}{\textbf{79.59\%}} &
        \multicolumn{1}{r|}{77.41\%} \\
        & 1f & O2 &
        \multicolumn{1}{r|}{10.45\%} &
        \multicolumn{1}{r|}{72.03\%} &
        \multicolumn{1}{r|}{57.45\%} &
        \multicolumn{1}{r|}{62.06\%} &
        \multicolumn{1}{r|}{70.31\%} &
        \multicolumn{1}{r|}{68.21\%} &
        \multicolumn{1}{r|}{\textbf{73.44\%}} &
        \multicolumn{1}{r|}{71.85\%} \\
        & 1g & O3 &
        \multicolumn{1}{r|}{45.72\%} &
        \multicolumn{1}{r|}{69.54\%} &
        \multicolumn{1}{r|}{75.64\%} &
        \multicolumn{1}{r|}{77.23\%} &
        \multicolumn{1}{r|}{77.72\%} &
        \multicolumn{1}{r|}{75.77\%} &
        \multicolumn{1}{r|}{\textbf{78.64\%}} &
        \multicolumn{1}{r|}{77.28\%} \\
        & 1h & O4 &
        \multicolumn{1}{r|}{69.16\%} &
        \multicolumn{1}{r|}{79.18\%} &
        \multicolumn{1}{r|}{79.84\%} &
        \multicolumn{1}{r|}{81.48\%} &
        \multicolumn{1}{r|}{80.81\%} &
        \multicolumn{1}{r|}{\textbf{81.58\%}} &
        \multicolumn{1}{r|}{80.81\%} &
        \multicolumn{1}{r|}{79.97\%} \\
      \hline
        \multirow{4}{*}{Food-101}
        & 1i & O1 &
        \multicolumn{1}{r|}{69.89\%} &
        \multicolumn{1}{r|}{69.84\%} &
        \multicolumn{1}{r|}{71.83\%} &
        \multicolumn{1}{r|}{71.57\%} &
        \multicolumn{1}{r|}{71.99\%} &
        \multicolumn{1}{r|}{72.32\%} &
        \multicolumn{1}{r|}{\textbf{72.46\%}} &
        \multicolumn{1}{r|}{71.45\%} \\
        & 1j & O2 &
        \multicolumn{1}{r|}{0\%} &
        \multicolumn{1}{r|}{55.41\%} &
        \multicolumn{1}{r|}{69.29\%} &
        \multicolumn{1}{r|}{71.23\%} &
        \multicolumn{1}{r|}{71.95\%} &
        \multicolumn{1}{r|}{\textbf{72.22\%}} &
        \multicolumn{1}{r|}{71.75\%} &
        \multicolumn{1}{r|}{70.73\%} \\
        & 1k & O3 &
        \multicolumn{1}{r|}{54.98\%} &
        \multicolumn{1}{r|}{61.27\%} &
        \multicolumn{1}{r|}{62.60\%} &
        \multicolumn{1}{r|}{63.10\%} &
        \multicolumn{1}{r|}{63.59\%} &
        \multicolumn{1}{r|}{63.32\%} &
        \multicolumn{1}{r|}{62.98\%} &
        \multicolumn{1}{r|}{\textbf{64.30\%}} \\
        & 1l & O4 &
        \multicolumn{1}{r|}{68.64\%} &
        \multicolumn{1}{r|}{69.55\%} &
        \multicolumn{1}{r|}{71.33\%} &
        \multicolumn{1}{r|}{72.28\%} &
        \multicolumn{1}{r|}{72.22\%} &
        \multicolumn{1}{r|}{\textbf{72.60\%}} &
        \multicolumn{1}{r|}{72.31\%} &
        \multicolumn{1}{r|}{71.33\%} \\
      \hline
    \end{tabularx}\label{tab:simple_dataset_optimizer_table}
  \end{table}
}

\setlength\tabcolsep{0in}
\begin{figure}[!htbp]
  \begin{tabular}{@{}lP{1.59in}P{1.59in}P{1.59in}P{1.59in}@{}}
    &
    \hskip 0.26in  O1 & \hskip 0.26in  O2 &
    \hskip 0.26in  O3 & \hskip 0.26in  O4 \\
    \raisebox{0.36in}{\rotatebox{90}{Imagenette}} &
    \includegraphics[width=1.59in]{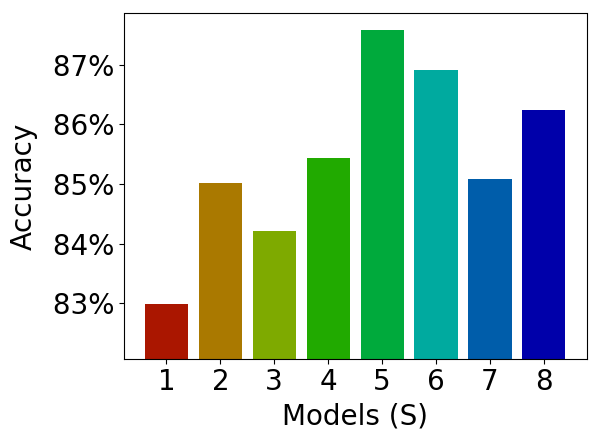} &
    \includegraphics[width=1.59in]{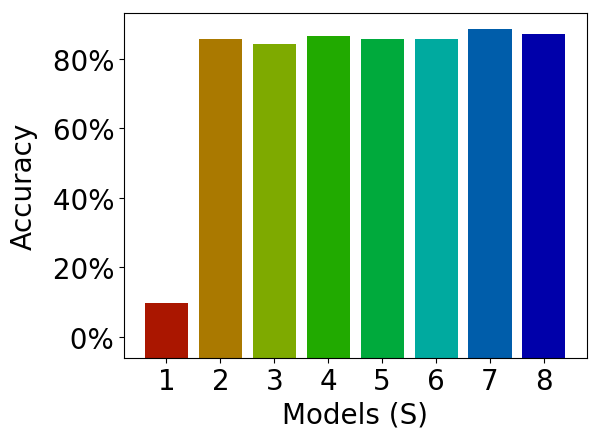} &
    \includegraphics[width=1.59in]{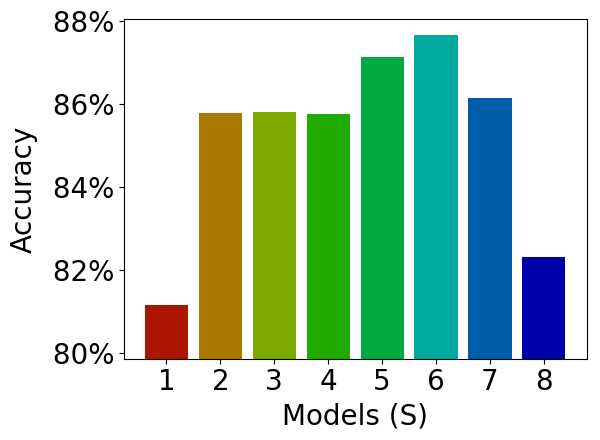} &
    \includegraphics[width=1.59in]{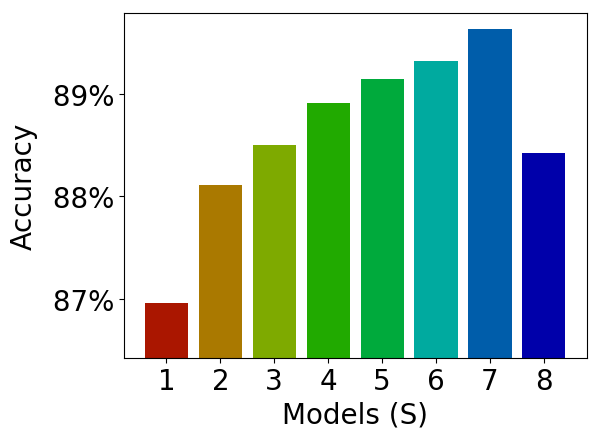} \\[-0.05in]
    &
    \hskip 0.26in  (1a) & \hskip 0.26in  (1b) &
    \hskip 0.26in  (1c) & \hskip 0.26in  (1d) \\[0.05in]
    \raisebox{0.35in}{\rotatebox{90}{Imagewoof}} &
    \includegraphics[width=1.59in]{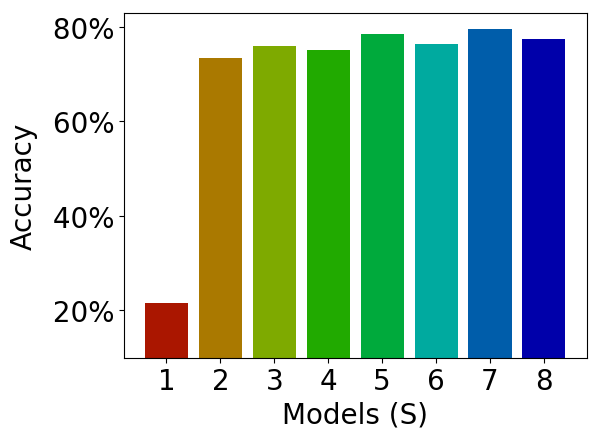} &
    \includegraphics[width=1.59in]{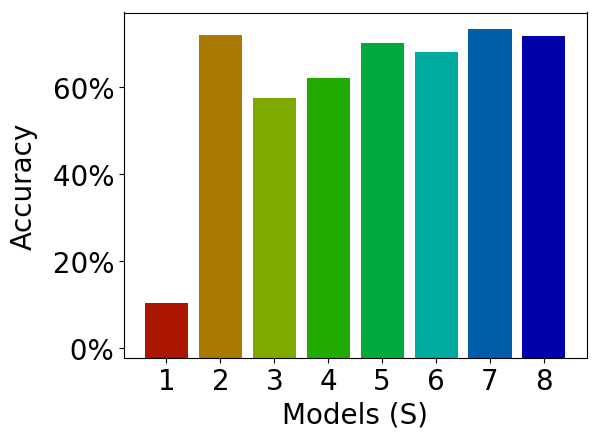} &
    \includegraphics[width=1.59in]{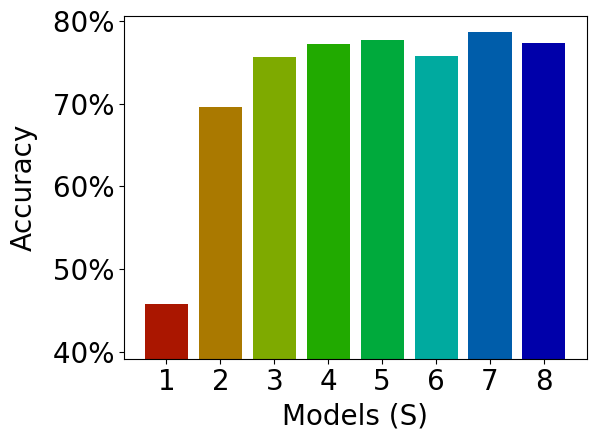} &
    \includegraphics[width=1.59in]{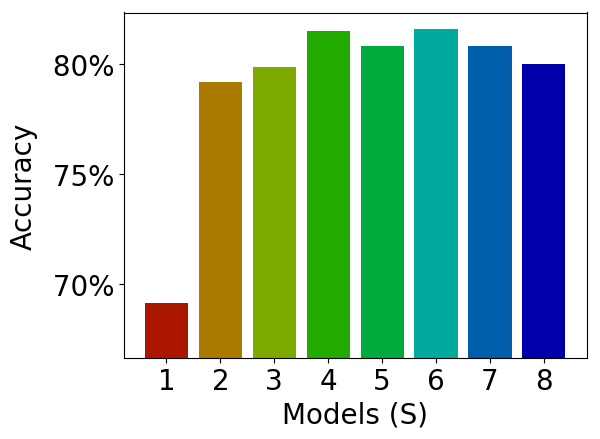} \\[-0.05in]
    &
    \hskip 0.26in  (1e) & \hskip 0.26in  (1f) &
    \hskip 0.26in  (1g) & \hskip 0.26in  (1h) \\[0.05in]
    \raisebox{0.4in}{\rotatebox{90}{Food-101}} &
    \includegraphics[width=1.59in]{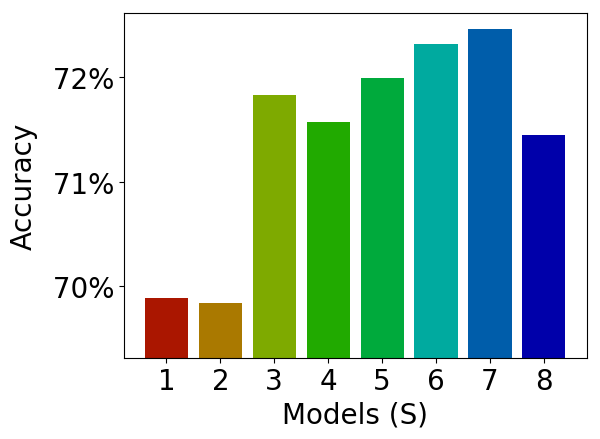} &
    \includegraphics[width=1.59in]{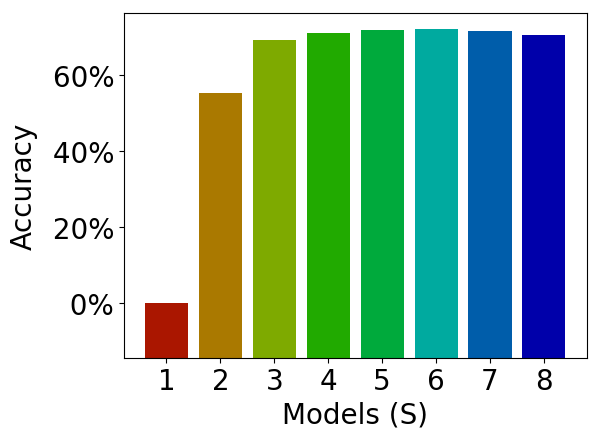} &
    \includegraphics[width=1.59in]{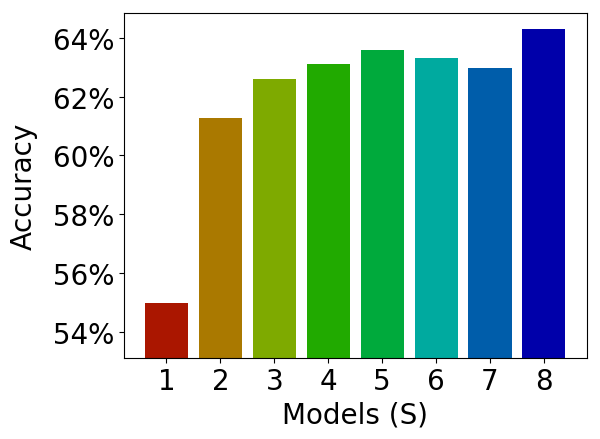} &
    \includegraphics[width=1.59in]{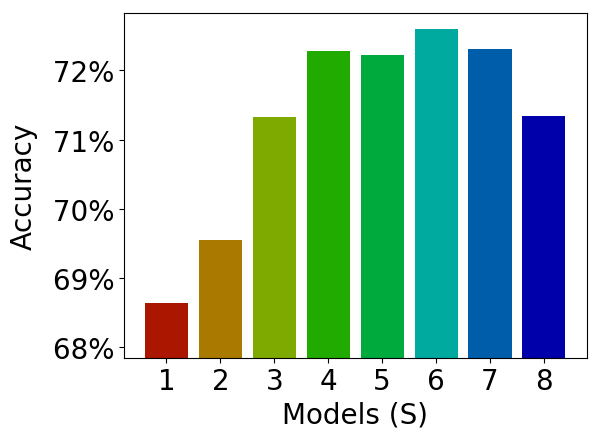} \\[-0.05in]
    &
    \hskip 0.26in  (1i) & \hskip 0.26in  (1j) &
    \hskip 0.26in  (1k) & \hskip 0.26in  (1l) \\
  \end{tabular}
  \caption{Classification accuracy on 4 simple CNN models, 3 different datasets, and 4 different optimization strategies.  Each chart is cross-referenced in \autoref{tab:simple_dataset_optimizer_table}}\label{fig:simple_dataset_optimizer_charts}
\end{figure}

{\renewcommand{\arraystretch}{1.4}%
  \setlength\tabcolsep{3pt}
  \begin{table}[!htbp]
    \caption{Classification accuracy on 8 Inception v3 models, 3 different datasets, and 4 different optimization strategies.  The first column in each row is a cross-referenc to the charts in \autoref{fig:inception_v3_dataset_optimizer_charts}.  The model parentheticals refer to the number of capsule dimensions.  For example I2 (8d) refers to model I2 which uses 8-dimensional capsules.}
    \begin{tabularx}{\textwidth}
      {|X|c|c|P{0.47in}|P{0.47in}|P{0.47in}|P{0.47in}|P{0.47in}|P{0.47in}|P{0.55in}|P{0.55in}@{}|}
      \hline
        \multirow{4}{*}{Dataset} & &
        \multirow{4}{*}{Optimizer} &
        \multicolumn{8}{c|}{Models (see \autoref{tab:inception_v3_models})} \\
        \cline{4-11} &
        \multirow{1}{*}{{\small \autoref{fig:inception_v3_dataset_optimizer_charts}}}   & &
        \multicolumn{1}{c|}{I1} &
        \multicolumn{1}{c|}{I2} &
        \multicolumn{1}{c|}{I3} &
        \multicolumn{1}{c|}{I4} &
        \multicolumn{1}{c|}{I5} &
        \multicolumn{1}{c|}{I6} &
        \multicolumn{1}{c|}{I7} &
        \multicolumn{1}{c|}{I8} \\
        & Chart & &
        \multicolumn{1}{c|}{} &
        \multicolumn{1}{c|}{(8d)} &
        \multicolumn{1}{c|}{(16d)} &
        \multicolumn{1}{c|}{(32d)} &
        \multicolumn{1}{c|}{(64d)} &
        \multicolumn{1}{c|}{(128d)} &
        \multicolumn{1}{c|}{(2,048d/} & 
        \multicolumn{1}{c|}{(64d/} \\   
        & & &
        \multicolumn{1}{c|}{} &
        \multicolumn{1}{c|}{} &
        \multicolumn{1}{c|}{} &
        \multicolumn{1}{c|}{} &
        \multicolumn{1}{c|}{} &
        \multicolumn{1}{c|}{} &
        \multicolumn{1}{c|}{\hspace{.05in}128d)} & 
        \multicolumn{1}{c|}{\hspace{.05in}25d)} \\ 
      \hline
        \multirow{4}{*}{Imagenette} & 2a & O1 &
        \multicolumn{1}{r|}{90.24\%} &
        \multicolumn{1}{r|}{89.50\%} &
        \multicolumn{1}{r|}{89.65\%} &
        \multicolumn{1}{r|}{89.60\%} &
        \multicolumn{1}{r|}{89.16\%} &
        \multicolumn{1}{r|}{\textbf{91.29\%}} &
        \multicolumn{1}{r|}{88.63\%} &
        \multicolumn{1}{r|}{86.12\%} \\
        & 2b & O2 &
        \multicolumn{1}{r|}{88.99\%} &
        \multicolumn{1}{r|}{88.73\%} &
        \multicolumn{1}{r|}{89.73\%} &
        \multicolumn{1}{r|}{90.37\%} &
        \multicolumn{1}{r|}{90.60\%} &
        \multicolumn{1}{r|}{\textbf{91.01\%}} &
        \multicolumn{1}{r|}{88.55\%} &
        \multicolumn{1}{r|}{85.43\%} \\
        & 2c & O3 &
        \multicolumn{1}{r|}{91.14\%} &
        \multicolumn{1}{r|}{92.09\%} &
        \multicolumn{1}{r|}{91.29\%} &
        \multicolumn{1}{r|}{91.03\%} &
        \multicolumn{1}{r|}{90.09\%} &
        \multicolumn{1}{r|}{89.98\%} &
        \multicolumn{1}{r|}{\textbf{92.16\%}} &
        \multicolumn{1}{r|}{91.29\%} \\
        & 2d & O4 &
        \multicolumn{1}{r|}{92.42\%} &
        \multicolumn{1}{r|}{91.73\%} &
        \multicolumn{1}{r|}{92.62\%} &
        \multicolumn{1}{r|}{\textbf{92.67\%}} &
        \multicolumn{1}{r|}{\textbf{92.67\%}} &
        \multicolumn{1}{r|}{\textbf{92.67\%}} &
        \multicolumn{1}{r|}{92.42\%} &
        \multicolumn{1}{r|}{92.47\%} \\
      \hline
        \multirow{4}{*}{Imagewoof} & 2e & O1 &
        \multicolumn{1}{r|}{79.71\%} &
        \multicolumn{1}{r|}{81.28\%} &
        \multicolumn{1}{r|}{81.17\%} &
        \multicolumn{1}{r|}{81.63\%} &
        \multicolumn{1}{r|}{80.48\%} &
        \multicolumn{1}{r|}{\textbf{84.14\%}} &
        \multicolumn{1}{r|}{74.67\%} &
        \multicolumn{1}{r|}{69.65\%} \\
        & 2f & O2 &
        \multicolumn{1}{r|}{79.48\%} &
        \multicolumn{1}{r|}{79.84\%} &
        \multicolumn{1}{r|}{80.02\%} &
        \multicolumn{1}{r|}{80.89\%} &
        \multicolumn{1}{r|}{81.89\%} &
        \multicolumn{1}{r|}{\textbf{83.38\%}} &
        \multicolumn{1}{r|}{78.02\%} &
        \multicolumn{1}{r|}{69.01\%} \\
        & 2g & O3 &
        \multicolumn{1}{r|}{85.99\%} &
        \multicolumn{1}{r|}{86.22\%} &
        \multicolumn{1}{r|}{85.63\%} &
        \multicolumn{1}{r|}{85.45\%} &
        \multicolumn{1}{r|}{86.24\%} &
        \multicolumn{1}{r|}{86.24\%} &
        \multicolumn{1}{r|}{\textbf{86.73\%}} &
        \multicolumn{1}{r|}{84.55\%} \\
        & 2h & O4 &
        \multicolumn{1}{r|}{84.73\%} &
        \multicolumn{1}{r|}{85.19\%} &
        \multicolumn{1}{r|}{84.81\%} &
        \multicolumn{1}{r|}{85.12\%} &
        \multicolumn{1}{r|}{84.89\%} &
        \multicolumn{1}{r|}{85.22\%} &
        \multicolumn{1}{r|}{\textbf{85.71\%}} &
        \multicolumn{1}{r|}{85.32\%} \\
      \hline
        \multirow{4}{*}{Food-101}
        & 2i & O1 &
        \multicolumn{1}{r|}{\textbf{80.00\%}} &
        \multicolumn{1}{r|}{77.86\%} &
        \multicolumn{1}{r|}{77.88\%} &
        \multicolumn{1}{r|}{78.20\%} &
        \multicolumn{1}{r|}{78.13\%} &
        \multicolumn{1}{r|}{78.32\%} &
        \multicolumn{1}{r|}{78.01\%} &
        \multicolumn{1}{r|}{76.05\%} \\
        & 2j & O2 &
        \multicolumn{1}{r|}{\textbf{82.52\%}} &
        \multicolumn{1}{r|}{80.51\%} &
        \multicolumn{1}{r|}{81.15\%} &
        \multicolumn{1}{r|}{81.76\%} &
        \multicolumn{1}{r|}{81.86\%} &
        \multicolumn{1}{r|}{81.03\%} &
        \multicolumn{1}{r|}{79.56\%} &
        \multicolumn{1}{r|}{76.99\%} \\
        & 2k & O3 &
        \multicolumn{1}{r|}{84.03\%} &
        \multicolumn{1}{r|}{84.43\%} &
        \multicolumn{1}{r|}{84.47\%} &
        \multicolumn{1}{r|}{84.33\%} &
        \multicolumn{1}{r|}{\textbf{84.49\%}} &
        \multicolumn{1}{r|}{84.06\%} &
        \multicolumn{1}{r|}{82.42\%} &
        \multicolumn{1}{r|}{82.13\%} \\
        & 2l & O4 &
        \multicolumn{1}{r|}{82.30\%} &
        \multicolumn{1}{r|}{82.97\%} &
        \multicolumn{1}{r|}{82.95\%} &
        \multicolumn{1}{r|}{82.98\%} &
        \multicolumn{1}{r|}{\textbf{83.22\%}} &
        \multicolumn{1}{r|}{82.92\%} &
        \multicolumn{1}{r|}{80.55\%} &
        \multicolumn{1}{r|}{79.43\%} \\
      \hline
    \end{tabularx}\label{tab:inception_v3_dataset_optimizer_table}
  \end{table}
}

\setlength\tabcolsep{0in}
\begin{figure}[!htbp]
  \begin{tabular}{@{}lP{1.59in}P{1.59in}P{1.59in}P{1.59in}@{}}
    &
    \hskip 0.26in  O1 & \hskip 0.26in  O2 &
    \hskip 0.26in  O3 & \hskip 0.26in  O4 \\
    \raisebox{0.36in}{\rotatebox{90}{Imagenette}} &
    \includegraphics[width=1.59in]{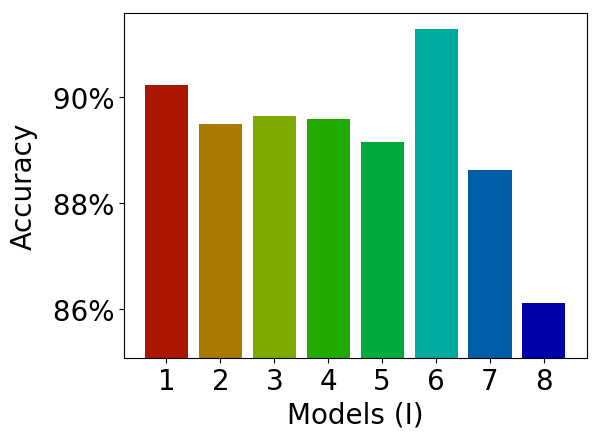} &
    \includegraphics[width=1.59in]{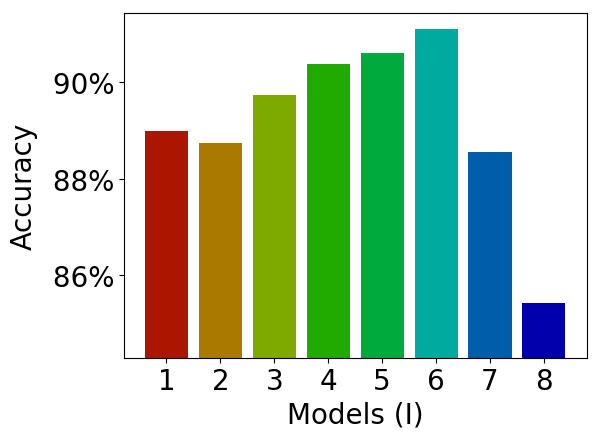} &
    \includegraphics[width=1.59in]{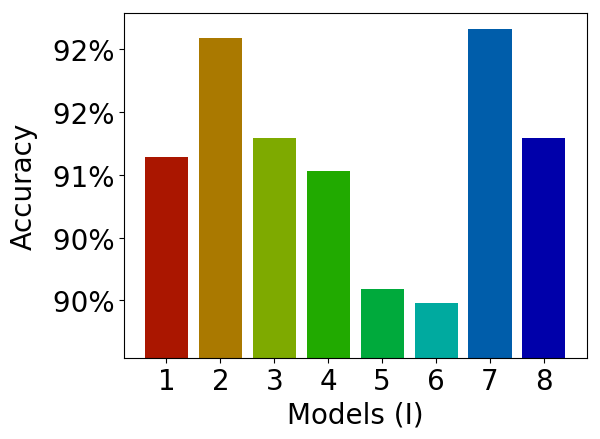} &
    \includegraphics[width=1.59in]{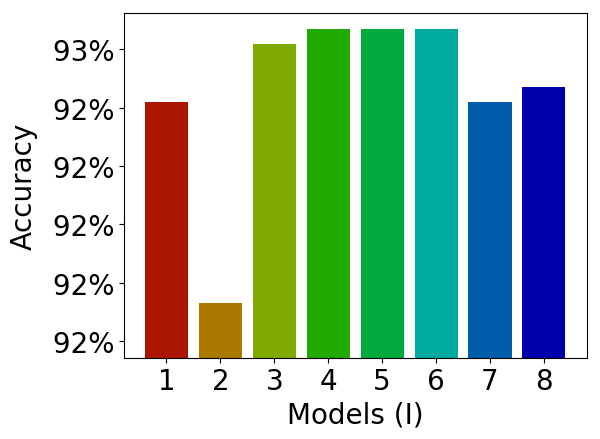} \\[-0.05in]
    &
    \hskip 0.26in  (2a) & \hskip 0.26in  (2b) &
    \hskip 0.26in  (2c) & \hskip 0.26in  (2d) \\[0.05in]
    \raisebox{0.35in}{\rotatebox{90}{Imagewoof}} &
    \includegraphics[width=1.59in]{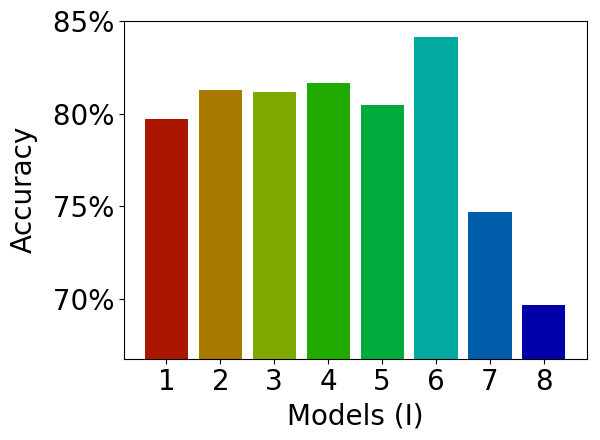} &
    \includegraphics[width=1.59in]{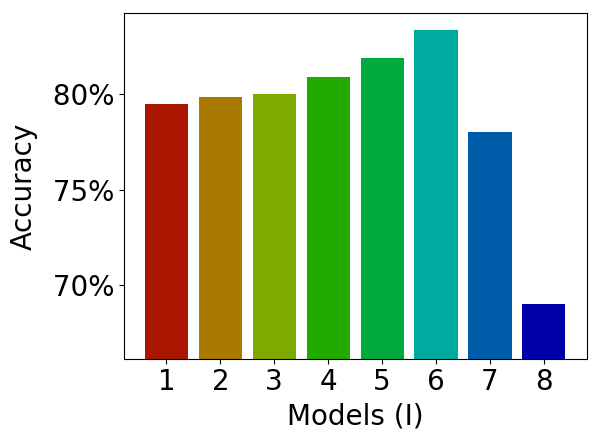} &
    \includegraphics[width=1.59in]{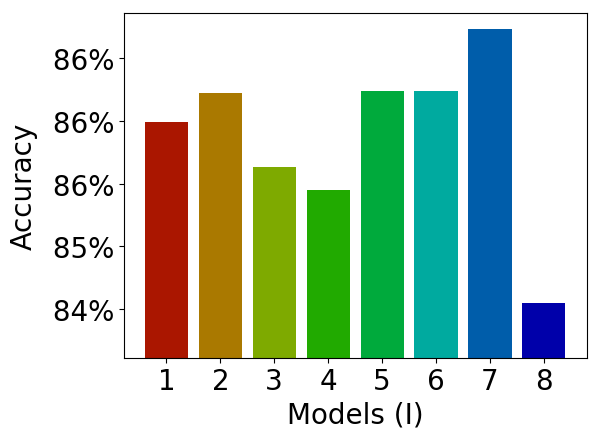} &
    \includegraphics[width=1.59in]{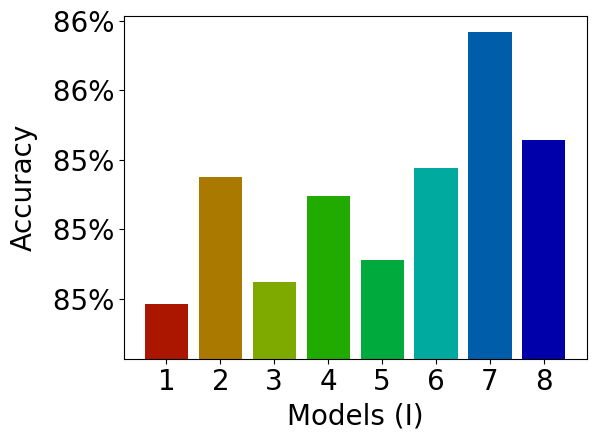} \\[-0.05in]
    &
    \hskip 0.26in  (2e) & \hskip 0.26in  (2f) &
    \hskip 0.26in  (2g) & \hskip 0.26in  (2h) \\[0.05in]
    \raisebox{0.4in}{\rotatebox{90}{Food-101}} &
    \includegraphics[width=1.59in]{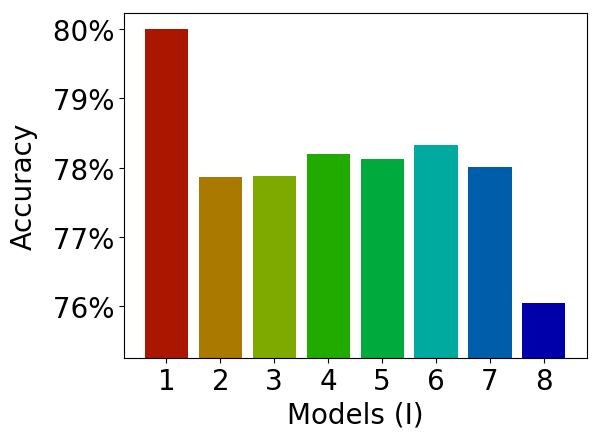} &
    \includegraphics[width=1.59in]{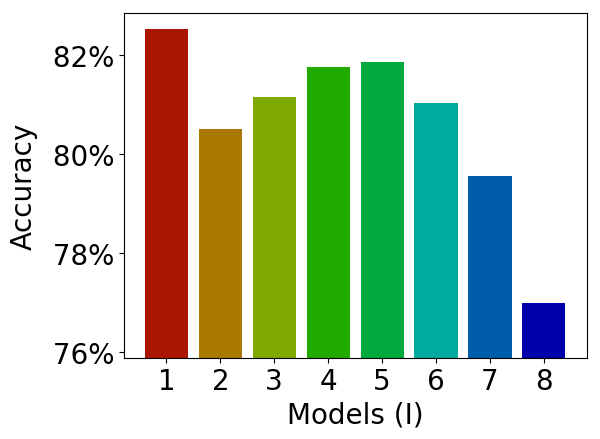} &
    \includegraphics[width=1.59in]{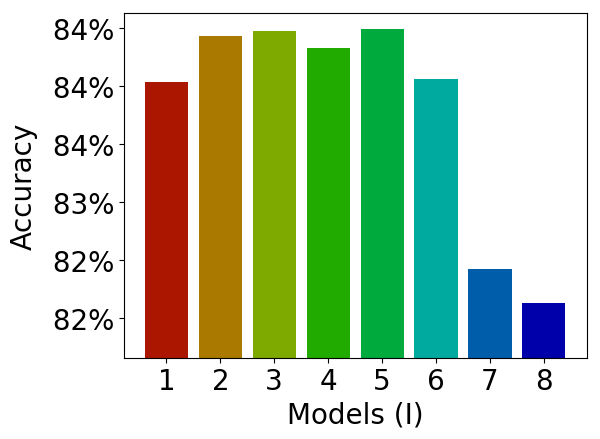} &
    \includegraphics[width=1.59in]{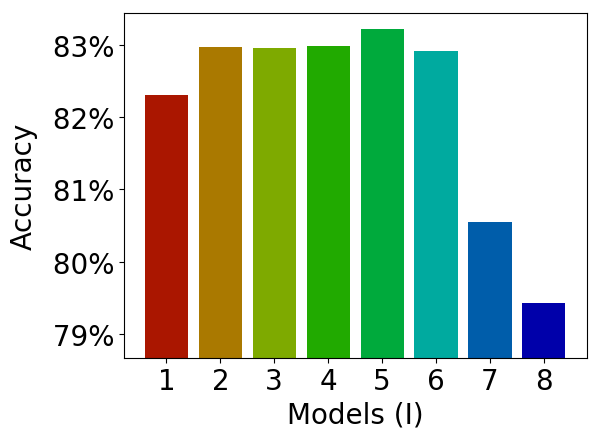} \\[-0.05in]
    &
    \hskip 0.26in  (2i) & \hskip 0.26in  (2j) &
    \hskip 0.26in  (2k) & \hskip 0.26in  (2l) \\
  \end{tabular}
  \caption{Classification accuracy on 8 Inception v3 models, 3 different datasets, and 4 different optimization strategies.  Each chart is cross-referenced in \autoref{tab:inception_v3_dataset_optimizer_table}}\label{fig:inception_v3_dataset_optimizer_charts}
\end{figure}

\section{Discussion}\label{sec:discussion}

\subsection{The Simple Monolithic CNN}\label{sec:discussion_simple_monolithic_cnn}

As can be seen in \autoref{tab:simple_dataset_optimizer_table} and \autoref{fig:simple_dataset_optimizer_charts}, models S2-S8, which used HVCs, wildly outperformed the baseline model S1 with all optimization strategies, on all three datasets tested.  The experiment for the baseline model S1 when using optimization strategy O2 was not able to learn to a better accuracy than random guessing on Imagenette and Imagewoof and actually ``learned'' to achieve an accuracy of 0\% on Food-101.  The average accuracy of the experiments of the baseline model S1 with all optimization strategies, on all three datasets, excluding those experiments where the model had not learned to an accuracy better than random guessing, is 64.55\%.  The average accuracy of the experiments for models S2-S8 with all optimization strategies, on all three datasets is 76.92\%.  This is a relative improvement of 19.16\%.  For all three datasets, the best performing experiments used optimization strategy O4, which was the Adam optimizer with an appropriately small \(\epsilon{}\) resulting in the intended adaptability along with a slowly decaying base learning rate.

6 out of 12 of the combinations of optimization strategy and dataset achieved their highest accuracy with model S7, which used the method that creates a single capsule from each distinct \(x\) and \(y\) coordinate of the feature maps.  4 out of 12 of the combinations achieved their highest accuracy with model S6, which used the method that creates 2 capsules from each distinct \(x\) and \(y\) coordinate of the feature maps.  This suggests that deriving 1 or 2 capsules for each distinct \(x\) and \(y\) coordinate of the feature maps is superior to deriving a higher number of capsules from each such \(x\) and \(y\) coordinate (models S2-S5) or deriving the capsules from entire, individual feature maps (model S8).

\subsection{Inception v3}\label{sec:discussion_inception_v3}

Optimization strategies O1 and O2 are the two optimization strategies published and used to train Inception v3 on ImageNet~\cite{Szegedy2015b}.  It would be understandable, yet naïve, to assume that these optimization strategies would be superior choices in general.  But as can be seen in \autoref{tab:inception_v3_dataset_optimizer_table} and \autoref{fig:inception_v3_dataset_optimizer_charts}, only occasionally did either strategy O1 or O2 outperform O3, and only once did O2 outperform O4.  This demonstrates that finely-tuned hyperparameters are finely-tuned, not just to the network architecture, but also to the data.  

The only times the baseline model I1 outperformed all capsule models I2-I8 was for the Food-101 dataset when using optimization strategies O1 and O2.  The best performing capsule model outperformed the baseline model I1 by an average of 1.32\% across all optimization strategies and datasets.

3 out of 12 of the combinations of optimization strategy and dataset achieved their highest accuracy with model I7.  This stands in contrast to the experiments on the simple monolithic CNN where twice as many combinations were superior for the analogous model S7.  The two architectures are too dissimilar to draw any firm conclusions, but we hypothesize that there two factors contributing to this.  First, creating a single capsule for each distinct \(x\) and \(y\) coordinate of all feature maps of the main output for Inception v3 results in 2,048 dimensional capsules (as the final set of feature maps is 2,048 in number) compared to only 256 dimensions for the capsules coming out of the final set of feature maps in the simple monolithic CNN.\@  Second, the presence of the auxiliary output stem in Inception v3.  5 out of 12 combinations achieved their highest accuracy with model I6 and 3 out of 12 with I5.  These results are less conclusive than those with the simple monolithic CNN and permit less firm conclusions.  However, these experiments do suggest that a single capsule to a small number of capsules for each distinct \(x\) and \(y\) coordinate of all feature maps is the superior choice.

\subsection{Optimization Strategy}\label{sec:discussion_optimizer}

For models S1-S8 and for all three datasets tested, optimization strategy O4 achieved the highest accuracy.  The second highest accuracy was achieved with strategy O1 twice and with O2 once.  For models I1-I8 and for all three datasets, optimization strategy O3 achieved the highest accuracy twice and O4 once.

With the Food-101 dataset, arguably the most difficult of the three datasets tested, baseline model S1 performed better with the quasi-adaptive optimization strategy O1 than with either of the truly adaptive strategies O3 or O4.  And yet, strategy O2 achieved a top accuracy of 0\% for this model.  O1 and O2 are the same optimization algorithm, but parameterized differently.  Further, these parameterizations were not ad-hoc, but rather parameterizations that are published along with the Inception v3 architecture and perform well on the ImageNet dataset with that architecture.  This underscores just how important hyperparameter choice can be and how closely related to both network structure and dataset it truly is.  This in turn underscores the relative utility of an adaptive gradient descent method that is less reliant on hyperparameter choice.

With adaptive gradient descent methods, there is a base learning rate \(\eta{}\) that is the same for all parameters and a separate per-parameter learning rate that is adapted based on previous gradient updates to that parameter.  The two are multiplied together to determine each parameter's actual update.  With the Adam optimizer, the suggested base learning rate \(\eta{}\) is \(0.001\) and the range of possible values for the per-parameter update are \(0\) to \(10^{10}\).  After being multiplied together, this gives a range of possible per-parameter updates of \(0\) to \(10^7\).  This is exactly what optimization strategy O3 uses for each parameter for the duration of the training.  Optimization strategy O4 starts with this range for each parameter, and then gradually decays the base learning rate \(\eta{}\) over the epochs of training such that the resultant per-parameter updates are eventually constrained to a range of \(0\) to \(10^4\).  This is similar to, but far less extreme (by four orders of magnitude) than the dampening effect caused by using a large \(\epsilon{}\) in the denominator of the per-parameter term of an adaptive gradient descent method (contra its intended purpose), as is the case with optimization strategies O1 and O2.  Further, when decaying the \textit{base} learning rate in the manner of optimization strategy O4, the dampening is applied gradually over time as the parameter values descend the loss landscape, rather than statically for the duration of training (as in the case of a large \(\epsilon{}\)).

Effectively, by allowing the learning rates of different parameters to change based on what has previously been learned, an adaptive gradient descent method attempts to achieve the goals of exploitation and exploration simultaneously.  Exploration is achieved by decoupling each parameter from a single learning rate and exploitation is achieved by the coupling of each parameters' own learning rate to what had previously been learned.  Using an adaptive gradient descent method with a large \(\epsilon{}\) greatly reduces the amount of per-parameter exploitation possible.  This shoulders the machine learning engineer with the task of choosing just the right hyperparameters to balance this small amount of variability in exploitation with the proper amount of exploration---the very thing adaptive gradient descent methods are meant to alleviate.  This is why, when using a large \(\epsilon{}\),  we choose to characterize them as \textit{quasi}-adaptive.  By using a truly adaptive gradient descent method (one with an appropriately small \(\epsilon{}\)) and then decaying the \textit{base} learning rate during training, the simultaneous explore/exploit nature of the method is preserved early in training and then slowly shifted to be more exploitative on average, but still allowing each parameter to have its own still rather large range of possible explore vs.\ exploit dispositions.

\section{Conclusion}\label{sec:conclusion}

The advent of convolutional layers led to considerable improvement in the performance of neural networks in image classification tasks as compared to networks composed entirely of fully connected layers~\cite{LeCun1998}.  This is correctly attributed to the convolutional layers' ability to extract localized features that are more complicated than a single pixel.  The feature extractors do this by assigning meaning to the spatial relationships among pixels that are close to each other.  Such meaning is absent when using fully connected layers.  As the term ``full connected'' implies, in fully connected layers every pixel is able to be associated with every other pixel without regard to their relative positions in the image.  Giving meaning to spatial relationships among the pixels can be understood as enforcing constraints upon which neurons are allowed to be associated with each other using trainable parameters.  Understood in this way, the success of convolutional neural networks can thus be understood as, in part, resulting from applying constraints on which neurons are allowed to affect other neurons in the next layer.

We interpret homogeneous vector capsules as performing a similar function, at the output stage of a convolutional neural network, as convolutional layers perform at the input stage.  In the traditional design of the classification stage of a CNN as depicted in \autoref{fig:fully_connected_before_output}, every neuron is able to adapt independently during backpropagation.  We hypothesize that this fact combined with the fact that adaptive gradient descent methods adapt independent learning rates for every parameter imparts two orders of adaptability---or stated another way, ``too much'' ``freedom'' (to adapt to the training data).  This would indeed result in overfitting and a generalization gap as has been observed when using adaptive gradient descent with CNNs.  By reshaping the output of the final convolutional layer into vectors and then connecting those vectors to a classification layer also composed of vectors, we are constraining groups of n-dimensional vectors of neurons to train together.

Thus, HVCs enable convolutional neural network researchers to:

\begin{enumerate}
\item Use adaptive gradient descent methods when training CNNs without experiencing a generalization gap.
\item Save time and compute cycles searching for the best learning rates and learning rate decay schedules to use to train their network with a non-adaptive gradient descent method and instead use an adaptive gradient descent method that does not require this fine-tuning.
\end{enumerate}

In general, we hypothesize that fully connected layers of scalar valued neurons are indeed ``misguided'' (as per Hinton et al.\ in~\cite{Hinton2011}).  Specifically, that using them after the convolutional layers in a CNN works against the goal of preserving meaning in spatial relationships within the features of an image.  The first layer of capsules in a pair of HVCs, groups outputs from the preceding convolutional layer together, preserving the spatial relationships that have been learned as meaningful.  By ``routing'' them to a second layer of capsules via trainable vectors, groups of capsules (the first layer of HVCs) that have preserved feature extractions from the convolutional layers are allowed to learn when they should be associated with each other to make a classification prediction (the second layer of HVCs).

In summary, our experimentation demonstrates that:

\begin{enumerate}
  \item Using HVCs on an advanced neural network architecture like Inception v3 increases the achievable accuracy by a small but significant margin.
  \item Using HVCs on a simple monolithic CNN increases the achievable accuracy massively.
  \item Deriving 1 or 2 capsules from each distinct \(x\) and \(y\) coordinate of all feature maps outperforms both deriving a larger number of capsules in the same manner and deriving capsules from entire, individual feature maps.
\end{enumerate}

\printbibliography{}

\bigskip

The code used for all experiments is publicly available on GitHub at: \href{https://github.com/AdamByerly/HVCsEnableAGD}{https://github.com/AdamByerly/HVCsEnableAGD}

\end{document}